\documentclass{article}

\usepackage[T1]{fontenc}

\usepackage{microtype}
\usepackage{graphicx}
\usepackage{subcaption}
\usepackage{booktabs} 
\usepackage{tabularx}
\usepackage{float}

\usepackage{hyperref}
\usepackage{multirow}
\usepackage[table]{xcolor}

\usepackage[preprint]{icml2026}

\usepackage{amsmath}
\usepackage{amssymb}
\usepackage{mathtools}
\usepackage{amsthm}
\usepackage{pifont}
\usepackage{algorithm}
\usepackage{algorithmic}
\usepackage{tikz}
\usetikzlibrary{shapes.geometric, arrows.meta, positioning, fit, backgrounds, shadows, calc}

\usepackage[capitalize,noabbrev]{cleveref}

\theoremstyle{plain}

\theoremstyle{definition}

\theoremstyle{remark}

\icmltitlerunning{DiagramNet: End-to-End Recognition Framework for System-Level Diagrams}

\begin{document}

\twocolumn[
  \icmltitle{DiagramNet: An End-to-End Recognition Framework and Dataset for
    Non-Standard System-Level Diagrams}

  \icmlsetsymbol{equal}{*}

  \begin{icmlauthorlist}
    \icmlauthor{Jincheng Lou}{sic}
    \icmlauthor{Ruohan Xu}{sic}
    \icmlauthor{Jiapeng Li}{xjtu}
    \icmlauthor{Junyin Pi}{thu}
    \icmlauthor{Runzhe Tao}{sic}
    \par\vspace{0.15em}
    \icmlauthor{Weijian Fan}{sic}
    \icmlauthor{Xiao Tan}{ssm}
    \icmlauthor{Guojie Luo}{cs}
    \icmlauthor{Yibo Lin}{sic,ieda,baic}
  \end{icmlauthorlist}

  \icmlaffiliation{sic}{School of IC, Peking University, Beijing, China}
  \icmlaffiliation{xjtu}{College of Artificial Intelligence, Xi'an Jiaotong University, Xi'an, China}
  \icmlaffiliation{thu}{Department of Precision Instruments, Tsinghua University, Beijing, China}
  \icmlaffiliation{ssm}{School of Software and Microelectronics, Peking University, Beijing, China}
  \icmlaffiliation{cs}{School of Computer Science, Peking University, Beijing, China}
  \icmlaffiliation{ieda}{Institute of EDA, Peking University, Beijing, China}
  \icmlaffiliation{baic}{Beijing Advanced Innovation Center for IC, Beijing, China}

  \icmlcorrespondingauthor{Jincheng Lou}{jinchenglou@stu.pku.edu.cn}
  \icmlcorrespondingauthor{Yibo Lin}{yibolin@pku.edu.cn}

  \icmlkeywords{Electronic Design Automation, Multimodal Learning, Vision-Language Models, Diagram Recognition, Circuit Diagram Understanding, DiagramNet}

  \vskip 0.3in
]


\printAffiliationsAndNotice{}  
\begin{abstract}
System-level diagrams encode the architectural blueprint of chip design, specifying module functions, dataflows, and interface protocols.
However, non-standardized symbols and the scarcity of structured training data hinder existing multimodal large language models (MLLMs) from recognizing these diagrams.
To address this gap, we introduce DiagramNet, the first multimodal dataset for system-level diagrams,
comprising \textbf{10,977} connection annotations and \textbf{15,515} chain-of-thought QA pairs across four tasks: Listing, Localization, Connection, and Circuit QA.
Building on this dataset, we propose a progressive training pipeline together with a decoupled multi-agent workflow that decomposes complex visual reasoning into Perception, Reasoning, and Knowledge stages.
On the DiagramNet benchmark, integrating our \textbf{3B}-parameter model with the proposed workflow surpasses the 2025 EDA Elite Challenge winner and outperforms GPT-5, Claude-Sonnet-4, and Gemini-2.5-Pro by over \textbf{2}$\times$ in end-to-end evaluation.
Notably, the workflow generalizes beyond our model, boosting Task~1 performance by \textbf{128.7}$\times$ for Gemini-2.5-Pro and \textbf{12.4}$\times$ for GPT-5.
Furthermore, with only \textbf{60} images for detector adaptation, the method transfers effectively to AMSBench, achieving zero-shot connectivity reasoning on par with GPT-5 and Claude-Sonnet-4 while surpassing the AMS state-of-the-art method Netlistify.
\end{abstract}
\begin{figure}[t]
    \centering
    \includegraphics[width=\columnwidth]{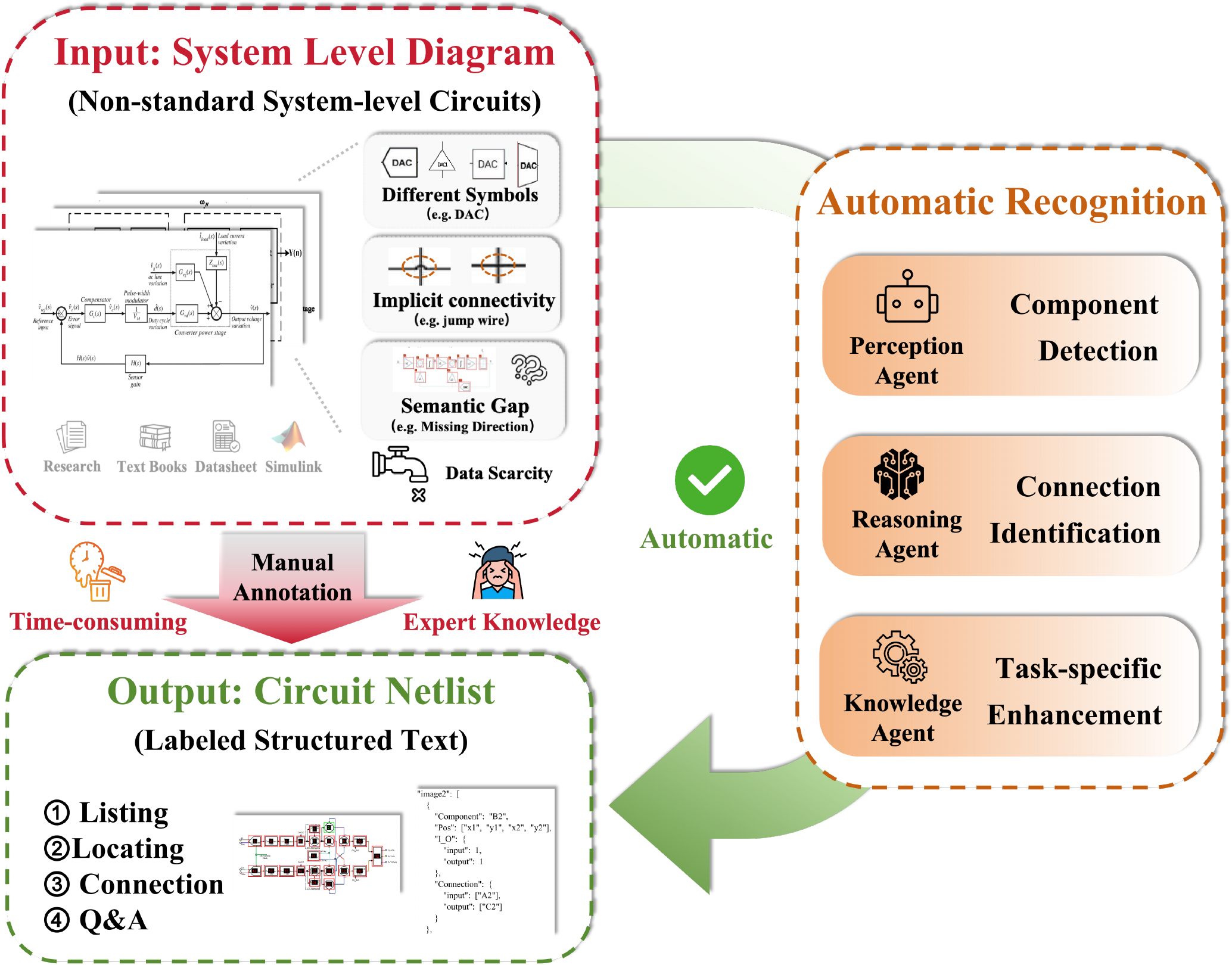}
\caption{Overview of the DiagramNet dataset and our end-to-end multi-agent workflow, which eliminates time-consuming manual recognition that requires domain expertise.}
    \label{fig:Intro}
\end{figure}
\section{Introduction}

AI techniques have advanced rapidly and are reshaping electronic design automation (EDA). Traditional methods include graph neural networks for chip placement~\cite{mirhoseini_graph_2021} and GPU-accelerated gradient-based optimization~\cite{9122053}. More recently, LLMs have enabled RTL code generation~\cite{liu_rtlcoder_2024,thakur_verigen_2023,lu_rtllm_2024}, syntax repair~\cite{tsai_rtlfixer_2024}, and domain-adapted engineering assistants~\cite{liu_chipnemo_2024}. Multi-agent systems further extend these capabilities: ChatEDA~\cite{wu_chateda_2024} orchestrates tool-augmented agents for end-to-end design automation, and VerilogCoder~\cite{ho_verilogcoder_2024} combines planning with waveform tracing for autonomous RTL development. Owing to their strong generalization, learning-based methods are becoming mainstream across the EDA pipeline.

Multimodal Large Language Models (MLLMs) have further enabled joint vision-language reasoning across diverse domains~\cite{openai2024gpt4technicalreport,geminiteam2025geminifamilyhighlycapable,bai2025qwen25vltechnicalreport}. Their capabilities, however, are fundamentally driven by high-quality datasets that provide structured supervision. In other visual domains, benchmarks such as AI2D~\cite{kembhavi_ai2d_2016}, ChartQA~\cite{masry_chartqa_2022}, and DocVQA~\cite{mathew_docvqa_2021} have driven substantial progress in structured visual reasoning. A similar benchmark could likewise advance structured understanding for circuit diagrams.

In EDA, existing benchmarks remain limited in scope. CircuitNet~\cite{chai_circuitnet_2023,jiang_circuitnet_2024} and VerilogEval~\cite{liu_verilogeval_2023} focus on digital backend prediction and RTL code generation, respectively. Multimodal circuit datasets are scarce and restricted to analog-mixed-signal (AMS) schematics: AMSNet~\cite{tao_amsnet_2024,shi_amsnet_2025} and Image2Net~\cite{xu_image2net_2025} provide schematic-netlist pairs, AMSBench~\cite{shi_amsbench_2025} offers evaluation tasks for AMS circuit understanding,
and Netlistify~\cite{huang_netlistify_nodate} constructs a 40k synthetic AMS schematic dataset.
Yet these efforts target standardized AMS schematics with fixed component libraries, and none provides the comprehensive annotations needed for any other level circuit diagrams.

Enabling MLLMs to understand circuit diagrams requires a dataset at a higher abstraction level. System-level diagrams occupy the top of the design hierarchy~\cite{weste2011cmos}, serving as the architectural blueprints of chip design. They capture how processors, memory controllers, PLLs, ADCs, and other functional blocks interconnect. This information then guides logic synthesis and physical layout.
Unlike transistor-level schematics or gate-level netlists, these diagrams use non-standardized domain-specific notations that vary across documents and organizations.
This visual diversity and the scarcity of public multimodal circuit data have limited progress on automated diagram understanding.
Most existing circuit datasets are restricted by NDAs or lack structured annotations. Moreover, manual annotation requires domain expertise and is highly time-consuming, which further exacerbates data scarcity, as illustrated in Figure~\ref{fig:Intro}.

To address these gaps, our contributions are as follows.
\begin{itemize}
    \item \textbf{A principled task formulation} decomposes system-level diagram recognition into four subtasks across three semantic levels: Listing and Localization at the perception level, Connection at the structure level, and Circuit QA at the semantic level. This hierarchical decomposition offers a general paradigm for structured visual reasoning, applicable beyond circuit diagrams to complex graph relationships and other technical illustrations with implicit connectivity.
    \item \textbf{DiagramNet}, the first multimodal dataset for system-level diagrams, contributes 10,977 connection pairs and 15,515 chain-of-thought QA samples with full annotations for all four subtasks. By bridging the data gap between general vision-language benchmarks and domain-specific EDA tasks, DiagramNet enables the community to study how MLLMs generalize to circuit domain.
    \item \textbf{A decoupled multi-agent workflow} decomposes complex visual reasoning into Perception, Reasoning, and Knowledge stages. This architecture not only achieves strong performance on DiagramNet and robust transfer to AMSBench with only 60 images for detector adaptation, but also provides a model-agnostic paradigm that boosts frontier MLLMs by up to 128$\times$.
    \item \textbf{A progressive training pipeline} combining supervised fine-tuning, topology-consistency reinforcement learning, and task-specific LoRA yields large gains on topology accuracy. This training recipe demonstrates how reinforcement learning with structured rewards can improve MLLM performance on tasks requiring precise relational reasoning.
\end{itemize}

\begin{figure*}[t]
    \centering
    \begin{subfigure}[b]{0.7\textwidth}
        \centering
        \includegraphics[height=5cm]{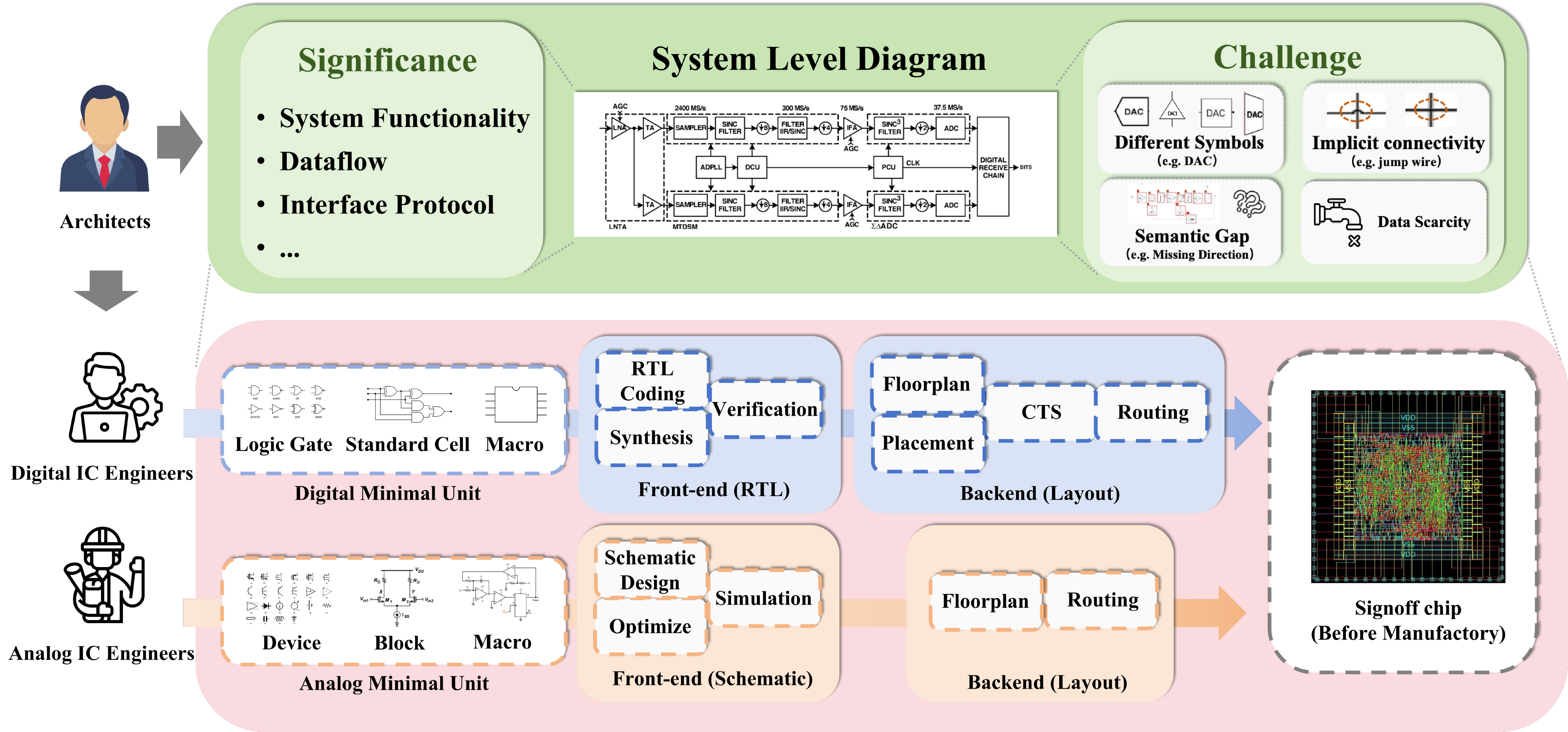}
        \caption{Relationship between system-level diagrams and the circuit design hierarchy}
        \label{fig:Prelim}
    \end{subfigure}
    \begin{subfigure}[b]{0.25\textwidth}
        \centering
        \includegraphics[height=5cm]{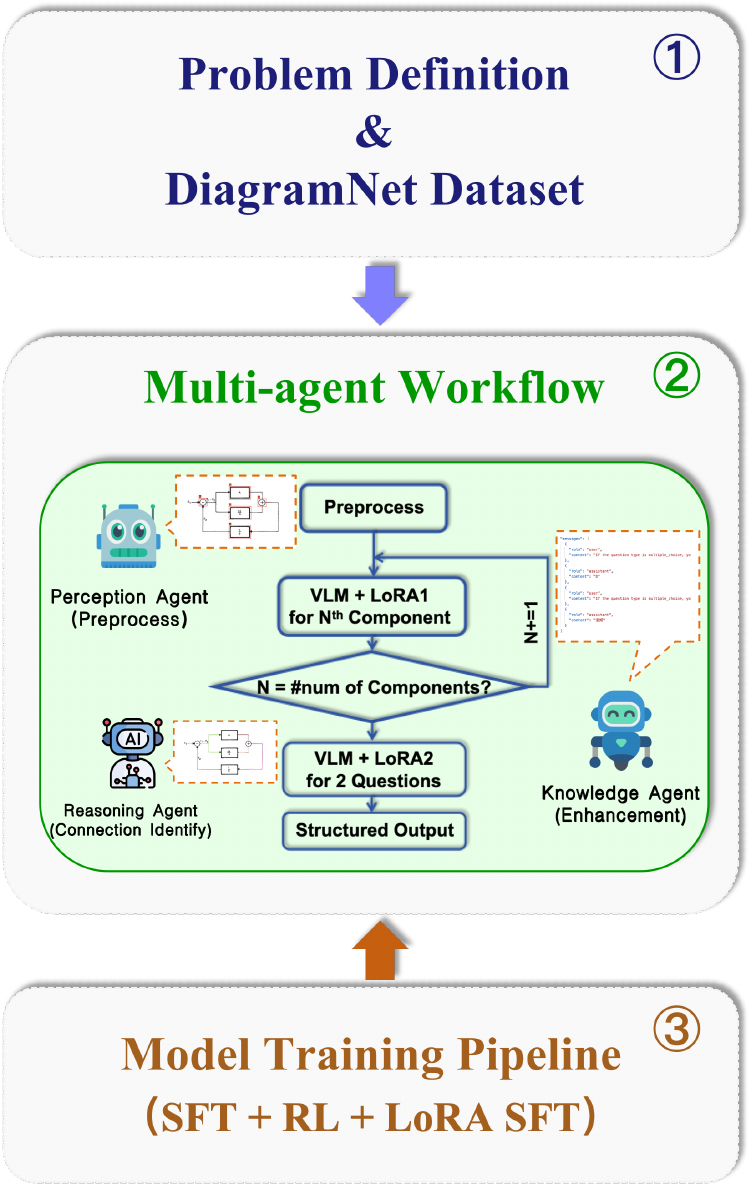}
        \caption{DiagramNet framework.}
        \label{fig:framework}
    \end{subfigure}
    \hspace*{0.02\textwidth}
    \caption{(a) System-level diagrams across three aspects: circuit type ({\colorbox{orange!80}{\textcolor{white}{\scriptsize\,A\,}}} analog, {\colorbox{blue!80}{\textcolor{white}{\scriptsize\,D\,}}} digital), design stage (front-end/back-end), and abstraction level (macro to device/gate).
    (b) The DiagramNet framework comprises three parts. The problem definition, DiagramNet dataset, and training pipeline together constitute an end-to-end recognition framework for system-level diagrams.}
    \label{fig:overview}
\end{figure*}


\section{Preliminary}

\textbf{Position in the design hierarchy.} We characterize system-level diagrams along three dimensions (Figure~\ref{fig:Prelim}).
First, by \textit{circuit type}: as indicated by the {\colorbox{orange!80}{\textcolor{white}{\scriptsize\,A\,}}} and {\colorbox{blue!80}{\textcolor{white}{\scriptsize\,D\,}}} regions, chips are divided into analog and digital categories, each handled by specialized IC engineers.
Second, by \textit{design stage}: following the arrows in the figure, the design flow proceeds from front-end to back-end. Although specific steps differ between analog and digital domains, the front-end generally addresses logical or functional implementation, while the back-end focuses on physical realization.
Third, by \textit{abstraction level}: at the leftmost end of the {\colorbox{orange!80}{\textcolor{white}{\scriptsize\,A\,}}} and {\colorbox{blue!80}{\textcolor{white}{\scriptsize\,D\,}}} regions, primitive building blocks differ by circuit type—devices for analog and gates for digital.
At the highest abstraction level, system-level diagrams represent both analog and digital chips, primarily defined by architects. They encode top-level architectural information including system functionality, dataflow, and interface protocols.

\textbf{Recognition challenges.}
System-level diagram understanding differs from AMS schematic recognition in four aspects.
\ding{182}~\textbf{Non-standardized symbols.} Unlike AMS circuits with a fixed set of component types, system-level diagrams have no upper limit on symbol categories. As shown in Figure~\ref{fig:Prelim}, even components with the same function can appear in very different visual forms, making template-based detection infeasible.
\ding{183}~\textbf{Implicit connectivity.} Unlike AMS circuits where devices have explicit input/output ports indicating connection direction, system-level diagrams from different sources may have directed or undirected connections. Jump wire labels and connection markers are also non-standardized.
\ding{184}~\textbf{Semantic gap.} Different system-level diagrams may represent circuits at different abstraction levels. They depict abstract architectures rather than real circuits, requiring deeper circuit understanding and reasoning.
\ding{185}~\textbf{Data scarcity.} No annotated system-level diagram dataset currently exists.

\section{Methodology}

This section presents the DiagramNet framework for recognizing system-level diagrams.
Figure~\ref{fig:framework} summarizes the overall pipeline that addresses heterogeneous symbols, implicit connections, and system-level semantics.
We first formalize the recognition problem into four subtasks in Section~\ref{sec:task-definition}.
We then describe the DiagramNet dataset with task-specific annotations in Section~\ref{sec:dataset}.
We design a multi-agent workflow with perception, reasoning, and knowledge agents in Section~\ref{sec:workflow}.
Finally, we present a 3-phase training pipeline that combines supervised fine-tuning, reinforcement learning, and low-rank adaptation in Section~\ref{sec:training}.

\subsection{DiagramNet Dataset}
\label{sec:dataset}

\begin{table*}[t]
\centering
\footnotesize
\renewcommand{\arraystretch}{1.15}
\setlength{\tabcolsep}{3pt}

\caption{Comparison of DiagramNet with AMS circuit datasets including AMSBench~\cite{shi_amsbench_2025}, Image2Net~\cite{xu_image2net_2025}, and Netlistify~\cite{huang_netlistify_nodate}. Li: Listing, Lo: Localization, C: Connection, QA: Circuit QA. $^\dagger$40k synthetic samples.}
\label{tab:dataset-comparison}

\begin{tabular}{l c c c c}
\toprule
& \textbf{DiagramNet} & \textbf{AMSBench} & \textbf{Image2Net} & \textbf{Netlistify} \\
\midrule
\multicolumn{5}{l}{\textit{Scope}} \\
\quad Abstraction level & System & AMS & AMS & AMS \\
\quad Supported tasks & Li, Lo, C, QA & Li, Lo, C, QA & Lo & Lo, C \\
\midrule
\multicolumn{5}{l}{\textit{Annotations}} \\
\quad \# Connections & 10,977 & 6,000 & 2,914 & 359$^\dagger$ \\
\quad \# QA pairs & 15,515 & 1,260 & --- & --- \\
\quad Coverage per diagram & All components & 1 component & All & All \\
\midrule
\multicolumn{5}{l}{\textit{Features}} \\
\quad Multimodal QA & \checkmark{} (+ CoT) & Text only & --- & --- \\
\quad Spatial annotation & Accurate Index & Approximate & --- & --- \\
\quad Includes method & \checkmark{} & --- & \checkmark{} & \checkmark{} \\
\quad Train / Eval split & \checkmark{} / \checkmark{} & --- / \checkmark{} & \checkmark{} / \checkmark{} & \checkmark{} / \checkmark{} \\
\bottomrule
\end{tabular}
\end{table*}

\subsubsection{Task Definition}
\label{sec:task-definition}

The first key contribution of this work is a principled decomposition of system-level diagram recognition into well-defined subtasks.
Unlike transistor-level schematics where components follow standard symbol libraries, system-level diagrams exhibit diverse visual styles and implicit connectivity.
Directly predicting the full circuit connection topology in one shot is difficult for current MLLMs.
We therefore define four subtasks:
\begin{equation}
\label{eq:task-formulation}
\left\{
\begin{aligned}
\textit{Listing:} \quad & f_{\text{list}}: I \rightarrow \mathcal{C} = \{c_1, \ldots, c_n\} \\
\textit{Localization:} \quad & f_{\text{loc}}: (I, c_i) \rightarrow b_i \in [0,1]^4 \\
\textit{Connection:} \quad & f_{\text{conn}}: (I, c_i, \mathcal{C}) \rightarrow \mathcal{T}_i \subseteq \mathcal{C} \\
\textit{Circuit QA:} \quad & f_{\text{qa}}: (I, q) \rightarrow (r, a)
\end{aligned}
\right.
\end{equation}
where $I$ denotes the input image, $\mathcal{C}$ is the component set ordered by row-major position index,
$b_i = (x, y, w, h)$ is a normalized bounding box, $\mathcal{T}_i \subseteq \mathcal{C} \setminus \{c_i\}$ is the set of output targets from $c_i$,
and $(r, a)$ denotes stepwise reasoning followed by the final answer.

\subsubsection{Dataset Construction and Composition}
DiagramNet is the first multimodal dataset for system-level diagram understanding.
The dataset contains 1,000 diagrams extracted from major chip design and computer architecture venues across multiple years, released as part of the 2025 EDA Elite Challenge~\cite{eda_elite_challenge_2025}. All images are publicly available conference and journal figures with no license restrictions. The dataset comprises 10,977 connection annotations and 15,515 chain-of-thought QA pairs.
We organize annotations into three semantic levels following the task hierarchy in Sec.~\ref{sec:task-definition}. All labels are produced through a semi-automated pipeline combining model pre-annotation with expert verification. The full dataset and code will be released upon publication.

\textbf{Perception level.} This level supports the Listing and Localization tasks.
Listing extracts component names from visual entities to establish the vocabulary for downstream tasks.
Localization provides spatial coordinates of component icons and serves as a reference for subsequent connection prediction.
After detection, components are reordered in row-major order from left to right and top to bottom to provide a consistent sequence for downstream reasoning. YOLOv11-nano detects component bounding boxes, Qwen2.5-VL extracts component names from cropped regions, and domain experts verify correctness. To handle diagrams with repeated component names, we assign unique identifiers via row-major ordering. The detector is trained as a single-class model to generalize across heterogeneous symbol styles.

\textbf{Structure level.} This level supports the Connection task.
Connection predicts only the output connections of a source component rather than all connections.
Gemini-2.5-Pro generates initial labels, which experts then verify and correct. The resulting connection topology annotations also provide ground truth for reinforcement learning rewards.

\textbf{Semantic level.} This level supports the Circuit QA task. We assemble multiple-choice questions at two levels: text-only questions adapted from circuit textbooks and image-conditioned questions generated from diagram content. The questions span seven domains including control systems, power electronics, RF communication, PLL design, data conversion, multi-rate DSP, and decision decoding. Each question includes stepwise reasoning to enable chain-of-thought training.

\textbf{Evaluation benchmark.} The evaluation benchmark follows Problem Two of the 2025 EDA Elite Challenge~\cite{eda_elite_challenge_2025}.
It contains 100 difficult diagrams with 100 Task~1 instances and 80 Task~2 instances.
Each image is manually verified to have no overlap with the training set.
No augmented or similar versions of training images appear in the test split.

\subsubsection{Comparison with prior work.}
As shown in Table~\ref{tab:dataset-comparison}, AMSBench, Image2Net, and Netlistify target standardized analog-mixed signal (AMS) schematics,
while DiagramNet focuses on system-level diagrams.
DiagramNet provides 1.8$\times$ more connection pairs than AMSBench (10,977 vs.\ 6,000) and 12.3$\times$ more QA samples (15,515 vs.\ 1,260), each with chain-of-thought.
Critically, DiagramNet annotates all components per diagram with relative positions in natural number ordering,
whereas AMSBench labels only one component per diagram with approximate locations.

\subsection{Multi-agent Workflow}
\label{sec:workflow}

\begin{figure*}[t]
    \centering
    \includegraphics[width=\textwidth]{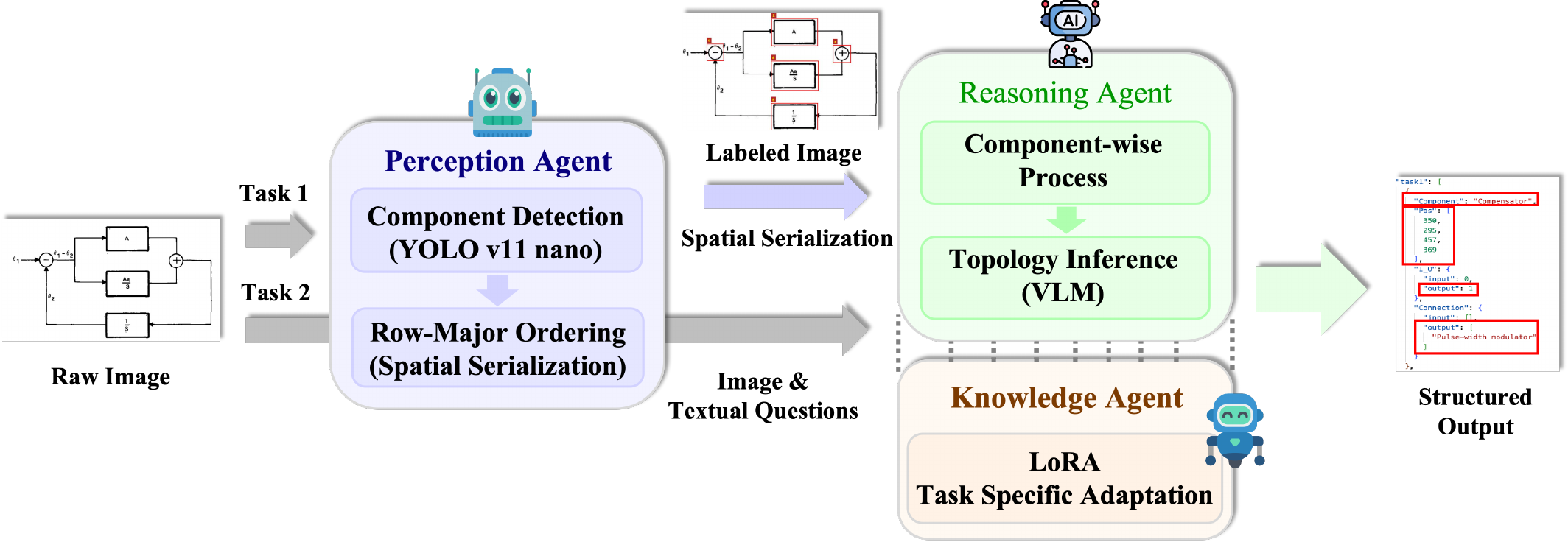}
    \caption{Architecture of the proposed multi-agent workflow. The \textbf{Perception Agent} detects components and applies row-major ordering to form a structured layout. The \textbf{Reasoning Agent} predicts directed connections per source component with a VLM backbone. The \textbf{Knowledge Agent} decides whether to activate task-specific LoRA adapters for Circuit QA and outputs the final answer.}
    \label{fig:multi_agent}
\end{figure*}

Recognition is decomposed into three specialized agents to keep each subproblem controllable.
Appendix~\ref{sec:app-case} and Figure~\ref{fig:mlmm-detect} show that end-to-end MLLMs suffer from visual grounding bottlenecks and spatial hallucinations on dense diagrams.
The term ``agent'' denotes an abstract functional role and allows the underlying model to be replaced independently.
The Perception Agent uses YOLOv11-nano and the Reasoning Agent uses a 3B VLM to enable low-cost transfer across diagram domains.
Figure~\ref{fig:multi_agent} shows the architecture of the multi-agent workflow and the inference procedure.
Following the task definitions in \S\ref{sec:task-definition}, we assign each subtask to a dedicated agent.
The Perception Agent outputs component locations and ordering indices.
The Reasoning Agent predicts component-wise topological connections.
The Knowledge Agent helps to enhance specific tasks.

\paragraph{Perception Agent.}
The perception agent converts raw pixels into a structured list of components with locations.
Vision-language models (VLMs) can detect objects but are costly and memory-heavy on large diagrams.
We instead use YOLO for efficient component localization.
To provide spatial structure for downstream reasoning,
we apply row-major ordering to the detected components.
This serialization removes ordering ambiguity and gives the reasoning agent a consistent positional prior.

\paragraph{Reasoning Agent.}
The reasoning agent is a vision-language model (VLM) that predicts connections between components.
Predicting the full circuit graph at once is challenging
because the output space is large and spatial constraints are complex.
We predict component-wise connections following the row-major order from the Perception Agent, from left to right and top to bottom.
The VLM sees the full diagram and a source component then predicts only its output connections.
In experiments, this approach improves connection identification accuracy, especially on dense diagrams with many crossings.

\paragraph{Knowledge Agent.}
The knowledge agent is responsible for loading task-specific LoRA weights into the VLM based on the tasks.
The purpose of introducing the knowledge agent is to improve performance on Circuit QA and to enable low-cost adaptation to new tasks.
This design enables efficient specialization to new circuit domains
while sharing one basic reasoning backbone.

\begin{algorithm}[t]
\caption{Multi-agent Workflow for System-Level Diagram Recognition}
\label{alg:workflow}
\begin{algorithmic}[1]
\REQUIRE System-level diagram $I$, query $Q$
\ENSURE Connectivity graph $\mathcal{G}$, answer $A$

\STATE \textsc{[Perception Agent]}
\STATE $\mathcal{B} \gets \textsc{Detect}(I)$ \COMMENT{component bounding boxes}
\STATE $\mathcal{B} \gets \textsc{SortRowMajor}(\mathcal{B})$ \COMMENT{spatial ordering}
\STATE $\mathcal{C} \gets \textsc{ExtractNames}(I, \mathcal{B})$ \COMMENT{component list}

\STATE
\STATE \textsc{[Reasoning Agent]}
\STATE $\mathcal{E} \gets \emptyset$
\FOR{each $c_i \in \mathcal{C}$}
    \STATE $\mathcal{T}_i \gets f_{\text{conn}}(I, c_i, \mathcal{C})$ \COMMENT{predict output targets}
    \STATE $\mathcal{E} \gets \mathcal{E} \cup \{(c_i, t) : t \in \mathcal{T}_i\}$
\ENDFOR
\STATE $\mathcal{G} \gets (\mathcal{C}, \mathcal{E})$ \COMMENT{directed topology graph}

\STATE
\STATE \textsc{[Knowledge Agent]}
\STATE $A \gets f_{\text{qa}}(I, Q; \theta_{\text{LoRA}})$ \COMMENT{domain reasoning}

\STATE \textbf{return} $\mathcal{G}, A$
\end{algorithmic}
\end{algorithm}
Algorithm~\ref{alg:workflow} summarizes the inference procedure.
The perception agent first detects and orders components.
The reasoning agent then iterates over each component to predict component-wise output connections.
Finally, the knowledge agent answers queries using the extracted topology or direct visual reasoning based on the question type.

\subsection{Model Training Pipeline}
\label{sec:training}

This section describes how we train the Reasoning Agent (VLM) and inject circuit connection and QA knowledge via LoRA.
The Perception Agent (YOLOv11-nano) is trained separately as a single-class detector. This step is lightweight and omitted here for brevity.
The training pipeline is divided into three phases and described in Sections~\ref{sec:training-sft} to \ref{sec:training-lora}.
Phase 1 uses multi-task supervised fine-tuning on all parameters.
Phase 2 applies reinforcement learning on hard samples.
Phase 3 uses LoRA for task-specific adaptation.

\subsubsection{Phase 1: Multi-task Supervised Fine-Tuning (SFT)}
\label{sec:training-sft}

General VLMs are pre-trained on natural images.
They focus on overall semantics, not precise logical connections.
To avoid cold-start issues, we initialize from a base model pre-trained on structured diagram tasks that share topological characteristics with circuit schematics.
Subsequent experiments~\ref{tab:overall-results} confirm that this choice yields better performance in understanding topological relationships.
This initialization gives the model a better starting point.
It avoids re-learning basic chart understanding.

We convert all four subtasks in Equation~\ref{eq:task-formulation} into instruction-tuning format.
The model takes image $X_v$ and instruction $X_t$ as input.
It generates every task's target answer sequence $Y$. We use cross-entropy loss:
\begin{equation}
\mathcal{L}_{\text{SFT}} = - \sum_{i=1}^{L} \log P(y_i | X_v, X_t, y_{<i}; \Theta)
\end{equation}
where $\Theta$ is model parameters and $L$ is sequence length.
We mix all four subtasks in each batch.
This balances learning across component recognition, localization, connection identification, and Circuit QA.

\subsubsection{Phase 2: Hard-Sample Reinforcement Learning with Compound Rewards}
\label{sec:training-rl}

Supervised fine-tuning works well on simple samples.
But it struggles with hard cases such as dense connections, crossing buses, or unusual layouts.
We use reinforcement learning to improve robustness on selected difficult samples.

\paragraph{Hard Sample Selection.}
We select difficult samples from DiagramNet using three criteria~\cite{Guo_2025,singh2024humandatascalingselftraining}.
\ding{182}~\textbf{Inference instability} captures cases where predictions vary across repeated decoding runs and yield low or inconsistent Pass@K scores.
\ding{183}~\textbf{Visual ambiguity} includes partially hidden or low resolution symbols and inconsistent diagram conventions such as unlabeled line crossings or reused icons.
\ding{184}~\textbf{High density connectivity} refers to large fan in and fan out with many crossings, which stresses topological consistency.

\textbf{Compound Reward Design.} We design rewards for each subtask. The total reward combines format validity ($R_{fmt}$) and prediction accuracy ($R_{acc}$):
\begin{equation}
R_{total} = \sum_{t \in \{loc, conn, qa, list\}} (\lambda_{f,t} R_{fmt}^{(t)} + \lambda_{a,t} R_{acc}^{(t)})
\end{equation}
The specific rewards are:

For \textbf{Localization}, IoU serves as the reward function:
\begin{equation}
R_{acc}^{(loc)} = \text{IoU}(bbox_{pred}, bbox_{gt})
\end{equation}
YOLO handles final inference, but RL training still includes localization to force the VLM's attention on correct component regions.

For \textbf{Connection}, F1 score plus a length penalty balances recall and precision:
\begin{equation}
R_{acc}^{(conn)} = \alpha \cdot F1(P, G) + (1-\alpha) \cdot R_{len}
\end{equation}
where $P$ and $G$ are predicted and ground truth sets. $R_{len}$ encourages matching the number of connections.

For \textbf{Listing}, F1 score plus LCS (Longest Common Subsequence) is applied:
\begin{equation}
R_{acc}^{(list)} = \beta_1 \cdot F1_{multi} + \beta_2 \cdot R_{len} + \beta_3 \cdot \frac{\text{LCS}(A, B)}{\max(|A|, |B|)}
\end{equation}
LCS rewards maintaining row-major order.

For \textbf{Circuit QA}, exact string matching is used:
\begin{equation}
R_{acc}^{(qa)} = \mathbb{I}(answer_{pred} == answer_{gt})
\end{equation}
These task-specific rewards define the RL objective and yield a robust base model.

\subsubsection{Phase 3: Final Refinement with LoRA}
\label{sec:training-lora}

After RL, the model has strong structured perception and logical reasoning.
To evaluate transferability to specific downstream tasks, LoRA (Low-Rank Adaptation) is applied for efficient fine-tuning.

The base model weights $W_0 \in \mathbb{R}^{d \times k}$ are frozen, and only low-rank matrices $B \in \mathbb{R}^{d \times r}$ and $A \in \mathbb{R}^{r \times k}$ are updated, where $r \ll \min(d, k)$. The forward pass is:
\begin{equation}
h = W_0 x + \Delta W x = W_0 x + \frac{\alpha}{r} B A x
\end{equation}
This adapts the output format such as Circuit QA without losing topology reasoning capability.
This shows our framework can serve as a general visual foundation for recognition any level circuits.

\section{Experimental Result}
This section demonstrates that the proposed method generalizes across benchmarks through comparison with diverse models and pipelines.
Experimental settings are described in \S\ref{sec:exp-setup}, followed by comparative results on the DiagramNet evaluation benchmark in \S\ref{sec:exp-compare} and ablation analyses in \S\ref{sec:exp-ablation}. Case studies are provided in Appendix~\ref{sec:app-case}.

\subsection{Experimental Settings}
\label{sec:exp-setup}
All experiments were conducted on NVIDIA A100-SXM4 (80GB) GPUs for reinforcement learning
and A40 GPUs for supervised fine-tuning, with LoRA adaptation tested on consumer-grade RTX 4090 GPUs.
The base model is \textbf{Qwen2.5-VL-3B-Hint-GRPO}~\cite{huang2025boostingmllmreasoningtextdebiased},
with weights pre-trained on AI2D-RST~\cite{Hiippala_2020} due to its topological similarity with circuit diagrams.
Training was implemented using LlamaFactory~\cite{zheng2024llamafactory} and Verl~\cite{sheng2024hybridflow}
with bfloat16 precision and DeepSpeed ZeRO-3 optimization.
For closed-source models, APIs were configured with temperature 0 and top\_p 0.9 to ensure reproducibility.

\begin{table*}[t]
\centering
\small
\renewcommand{\arraystretch}{1.15}
\setlength{\tabcolsep}{5pt}

\caption{DiagramNet benchmark results. Best in \textbf{bold}, second \underline{underlined}.}
\label{tab:overall-results}

\begin{tabular}{l l c c c c c c}
\toprule
& \textbf{Method} & \textbf{S1} & \textbf{S2} & \textbf{S3} & \textbf{Task~1} & \textbf{Task~2} & \textbf{Overall} \\
\midrule
\textit{Ours}
& DiagramNet-3B (w. Workflow) & \textbf{0.988} & \textbf{0.828} & \underline{0.735} & \underline{0.855} & 0.395 & \textbf{0.671} \\
\midrule
\multirow{2}{*}{\textit{Domain}}
& EDA Elite Winner & 0.984 &  \underline{0.787} & \textbf{0.777} & \textbf{0.862} & 0.370 & \underline{0.665} \\
& Netlistify~\cite{huang_netlistify_nodate} & \underline{0.986} & 0.239 & 0.150 & 0.502 & --- & --- \\
\midrule
\multirow{2}{*}{\textit{Open-source}}
& Qwen2.5-VL-3B~\cite{bai2025qwen25vltechnicalreport} & 0.111 & 0.038 & 0.014 & 0.058 & 0.360 & 0.179 \\
& Hint-GRPO~\cite{huang2025boostingmllmreasoningtextdebiased} & 0.153 & 0.062 & 0.025 & 0.083 & 0.355 & 0.192 \\
\midrule
\multirow{3}{*}{\textit{Commercial}}
& Gemini-2.5-Pro (E2E) & 0.008 & 0.005 & 0.008 & 0.007 & \underline{0.685} & 0.278 \\
& GPT-5 (E2E) & 0.085 & 0.068 & 0.029 & 0.059 & \textbf{0.730} & 0.327 \\
& Claude-Sonnet-4 (E2E) & 0.477 & 0.337 & 0.265 & 0.364 & 0.445 & 0.397 \\
\bottomrule
\end{tabular}
\end{table*}



\begin{table*}[t]
\centering
\small
\renewcommand{\arraystretch}{1.15}
\setlength{\tabcolsep}{5pt}

\caption{Multi-agent Workflow effect. Gain is vs. end-to-end inference. Best in \textbf{bold}.}
\label{tab:workflow-compare}

\begin{tabular}{l l c c c c c c c}
\toprule
& \textbf{Method} & \textbf{S1} & \textbf{S2} & \textbf{S3} & \textbf{Task~1} & \textbf{Task~2} & \textbf{Overall} & \textbf{Gain (T1)} \\
\midrule
\textit{Ours}
& DiagramNet-3B & \textbf{0.988} & \textbf{0.828} & \textbf{0.735} & \textbf{0.855} & 0.395 & \textbf{0.671} & — \\
\midrule
\multirow{2}{*}{\textit{Open-source}}
& Qwen2.5-VL-3B~\cite{bai2025qwen25vltechnicalreport} & 0.988 & 0.375 & 0.203 & 0.552 & 0.365 & 0.447 & 9.5$\times$ \\
& Hint-GRPO~\cite{huang2025boostingmllmreasoningtextdebiased} & 0.988 & 0.454 & 0.258 & 0.589 & 0.355 & 0.496 & 7.1$\times$ \\
\midrule
\multirow{3}{*}{\textit{Commercial}}
& Gemini-2.5-Pro & 0.986 & 0.870 & 0.832 & 0.901 & 0.685 & 0.815 & 128.7$\times$ \\
& GPT-5 & 0.986 & 0.582 & 0.555 & 0.733 & \textbf{0.705} & 0.722 & 12.4$\times$ \\
& Claude-Sonnet-4 & 0.986 & 0.486 & 0.314 & 0.618 & 0.450 & 0.551 & 1.7$\times$ \\
\bottomrule
\end{tabular}
\end{table*}

\subsection{Evaluation on DiagramNet benchmark}
\label{sec:exp-compare}

The evaluation is conducted on the DiagramNet benchmark from Problem Two of the 2025 EDA Elite Challenge~\cite{eda_elite_challenge_2025}.
Task~1 covers component detection (S1), output count (S2), and connection identification (S3).
Task~2 covers Circuit QA.
F1 scores are reported for all metrics.
The overall score follows the official formula:
\begin{equation}
\label{eq:overall-score}
\begin{aligned}
Score_{\text{Task1}} &= 0.4\,S_1 + 0.2\,S_2 + 0.4\,S_3, \\
Score_{\text{overall}} &= 0.6\,Score_{\text{Task1}} + 0.4\,Score_{\text{Task2}}.
\end{aligned}
\end{equation}

Table~\ref{tab:overall-results} reports the main results. DiagramNet-3B achieves an overall score of 0.671, ranking first among all methods.

\textbf{Comparison with domain-specific methods.} DiagramNet-3B outperforms EDA Elite Winner by \textbf{4.1\%}
on S2 and matches its S1 performance.
Netlistify, the AMS circuit recognition SOTA, achieves only 0.150 on S3, trailing DiagramNet-3B by \textbf{4.9$\times$}.
Even with our YOLO detector, its poor S3 performance indicates weak generalization from standardized AMS schematics to system-level diagrams.
The detailed analysis is provided in Appendix~\ref{sec:app-case}.

\textbf{Comparison with open-source MLLMs.}
Relative to Qwen2.5-VL-3B, DiagramNet-3B improves S2 by \textbf{21.8$\times$} and S3 by \textbf{52.5$\times$}.
Compared to Hint-GRPO, the gains remain substantial at \textbf{13.4$\times$} on S2 and \textbf{29.4$\times$} on S3.
These results demonstrate the effectiveness of the proposed training pipeline.
A detailed phase-wise analysis is provided in Section~\ref{sec:abla-training}.

\textbf{Comparison with commercial MLLMs.} In end-to-end mode, commercial models struggle on Task~1:
Gemini-2.5-Pro and GPT-5 score below 0.1, while Claude-Sonnet-4 reaches 0.364. Despite detailed prompts with output format guidance,
Gemini-2.5-Pro and GPT-5 fail to accurately recognize components, leading to cascading errors in downstream subtasks.
A detailed analysis is provided in Appendix~\ref{sec:app-case}.
DiagramNet-3B surpasses Claude-Sonnet-4 by \textbf{2.3$\times$} on Task~1. On Task~2, GPT-5 leads at 0.730,
exceeding DiagramNet-3B by 1.8$\times$.
This gap reflects the broader domain knowledge of commercial models from large-scale pretraining.
The analysis of this issue is shown in Section~\ref{sec:abla-training}.
Nevertheless, DiagramNet-3B achieves the best overall score in DiagramNet benchmark.
\subsection{Ablation Study}
\label{sec:exp-ablation}

\subsubsection{Validation of Multi-agent Workflow}

Table~\ref{tab:workflow-compare} shows that the multi-agent workflow provides substantial gains on Task~1 across all models.
On Task~1, the workflow boosts Gemini-2.5-Pro by \textbf{128.7}$\times$, GPT-5 by \textbf{12.4}$\times$,
and Claude-Sonnet-4 by \textbf{1.7}$\times$. It also improves the open-source baselines by \textbf{9.5}$\times$
and \textbf{7.1}$\times$. These results demonstrate that the workflow generalizes well and benefits models of varying capacity.
Notably, Gemini-2.5-Pro with our workflow achieves the highest Task~1 score of 0.901, surpassing DiagramNet-3B.
This shows the workflow is not tied to our model and can boost other VLMs.
DiagramNet-3B also outperforms GPT-5 and Claude-Sonnet-4 on Task~1, indicating strong standalone performance.
The improvement on Task~2 is limited because LoRA weights do not transfer across models.

\subsubsection{Validation of the training pipeline}
\label{sec:abla-training}

\begin{table*}[!t]
\centering
\small
\renewcommand{\arraystretch}{1.15}
\setlength{\tabcolsep}{6pt}

\caption{Ablation study on different training phases. Each row adds one phase to the previous configuration. S1 is fixed at 0.988 (handled by YOLOv11-nano) and omitted for clarity.}
\label{tab:stage-ablation}

\begin{tabular*}{\textwidth}{@{\extracolsep{\fill}}l*{5}{c}}
\toprule
\textbf{Training Phase} &
\textbf{S2} & \textbf{S3} & \textbf{Task~1} & \textbf{Task~2} & \textbf{Overall} \\
\midrule

Qwen2.5-VL-3B (base)        & 0.375 & 0.203 & 0.552 & 0.365 & 0.447 \\
\quad + Hint-GRPO           & 0.454 & 0.258 & 0.589 & 0.355 & 0.496 \\
\quad\quad + SFT            & 0.736 & 0.594 & 0.780 & 0.355 & 0.610 \\
\quad\quad\quad + RL        & 0.807 & 0.725 & 0.847 & 0.355 & 0.650 \\
\quad\quad\quad\quad + LoRA & \textbf{0.828} & \textbf{0.735} & \textbf{0.855} & \textbf{0.395} & \textbf{0.671} \\

\bottomrule
\end{tabular*}
\end{table*}

%
%
Table~\ref{tab:stage-ablation} summarizes the phase-wise effects.
Comparing Qwen2.5-VL-3B with Hint-GRPO shows that training on AI2D-RST improves Task~1 but degrades Task~2.
This reveals that 3B models are easy to forget during fine-tuning.
Optimizing one task can degrade performance on another.
To address this issue, we train Task~1 and Task~2 jointly by combining their data in each training phase.
This joint training prevents Task~2 from dropping while Task~1 improves, so preserving Task~2 accuracy is itself a success.
The final model improves S2 by \textbf{2.2}$\times$, S3 by \textbf{3.6}$\times$, overall score by \textbf{22.4\%}, and Task~2 by \textbf{3\%} over Qwen2.5-VL-3B. S1 remains unchanged because it is handled by a fixed YOLOv11-nano detector.
These results validate the effectiveness of each phase in the training pipeline.

\subsubsection{Validation of Task Transferability}

This ablation validates the generalization capability of DiagramNet-3B and its genuine understanding of circuit connectivity.
Transferability is evaluated by stratified sampling on AMSBench~\cite{shi_amsbench_2025} with a ratio of \textbf{7:2:1} over easy, medium, and hard cases, using \textbf{30} cases per task and a fixed seed of \textbf{324964}.
YOLOv11-nano is retrained with only \textbf{60} images from AMSBench to adapt to AMS components.
Because these images are used for detector training,
component detection tasks are excluded, and the evaluation focuses on two connection tasks and Textual QA to test connectivity reasoning and domain knowledge.

Table~\ref{tab:AMSBench_result} reports zero-shot transfer results on AMSBench.
DiagramNet-3B achieves 0.500/0.667/0.900 on Connection Identification, Connection Judgement, and TQA.
Percentages denote relative change to the compared method.
On Connection Identification, it improves over the 2025 EDA Elite winner at 0.133 by 36.7\%, exceeds Netlistify at 0.433 by 6.7\%, and outperforms Hint-GRPO at 0.233 by 26.7\%.
On Connection Judgement, it ties Netlistify and Hint-GRPO at 0.667, improves over the EDA winner at 0.567 by 10\%, and trails GPT-5 at 0.700 by 0.3\%.
On TQA, it ties GPT-5 at 0.900, exceeds Hint-GRPO and Gemini-2.5-Pro at 0.867 by 3.3\%.
\begin{table}[t]
\centering
\small
\renewcommand{\arraystretch}{1.15}
\setlength{\tabcolsep}{5pt}

\definecolor{grayblock}{RGB}{240,240,240}

\caption{Zero-shot connectivity reasoning on AMSBench. Open-source models are evaluated with the multi-agent workflow, while commercial APIs (gray rows) are evaluated end-to-end. Best in \textbf{bold}, second \underline{underlined}.}
\label{tab:AMSBench_result}

\begin{tabularx}{\columnwidth}{l*{3}{>{\centering\arraybackslash}X}}
\toprule
\textbf{Method} & \textbf{Conn. Ident.} & \textbf{Conn. Judg.} & \textbf{TQA} \\
\midrule
DiagramNet-3B                       & \underline{0.500} & \underline{0.667} & 0.900  \\
EDA Elite Winner & 0.133             & 0.567          & \underline{0.930} \\
Netlistify~\cite{huang_netlistify_nodate} & 0.433       & 0.667          & —              \\
Hint-GRPO~\cite{huang2025boostingmllmreasoningtextdebiased} & 0.233 & 0.667 & 0.867        \\
\midrule
\rowcolor{grayblock}
Gemini-2.5-Pro                      & \textbf{0.670}    & \textbf{1.000} & 0.867          \\
\rowcolor{grayblock}
GPT-5                               & 0.530             & 0.700          & 0.900          \\
\rowcolor{grayblock}
Claude-Sonnet-4                     & 0.200             & 0.800          & \textbf{0.933}          \\
\bottomrule
\end{tabularx}
\end{table}

\section{Conclusion}

We presented DiagramNet, the first multimodal dataset and recognition framework for system-level diagrams.
The dataset contributes 10,977 connection annotations and 15,515 chain-of-thought QA pairs,
addressing a critical data gap in circuit understanding.
Our hybrid training pipeline combines supervised fine-tuning, topology-consistency reinforcement learning,
and task-specific LoRA. It improves S2 by 21.86$\times$ and S3 by 52.57$\times$ over Qwen2.5-VL-3B,
achieving an overall score of 0.671 that surpasses EDA Elite Winner by 22.4\%.
The multi-agent workflow generalizes across model scales, yielding 128.7$\times$ Task~1 improvement for Gemini-2.5-Pro and enabling competitive zero-shot transfer to AMSBench.




\section*{Software and Data}

Due to submission file size limits, dataset images and model weights are not included. The released scripts and annotation files are available at: \url{https://anonymous.4open.science/r/DiagramNet-1727/}. Upon acceptance, the full dataset, models, and code will be released.




\section*{Impact Statement}
This work advances multimodal AI for electronic design automation. 
DiagramNet provides the first public benchmark for system-level diagram recognition, 
addressing data scarcity and enabling reproducible evaluation. 
The multi-agent workflow offers a reusable paradigm for other level circuit diagrams such as PCB schematics and architectural blueprints. 
Competitive cross-domain results with only 60 images for detector adaptation lower the barrier for applying MLLMs where labeled data is scarce. 
Potential risks include exposure of proprietary designs and over-reliance on automation in safety-critical flows. 
We recommend human verification in high-stakes contexts.

\newpage
\appendix
\onecolumn
\section{Task Definitions and Prompts}
\label{sec:app-prompts}

\begin{figure}[ht]
    \centering
    \begin{subfigure}[t]{0.32\textwidth}
        \centering
        \includegraphics[width=\linewidth]{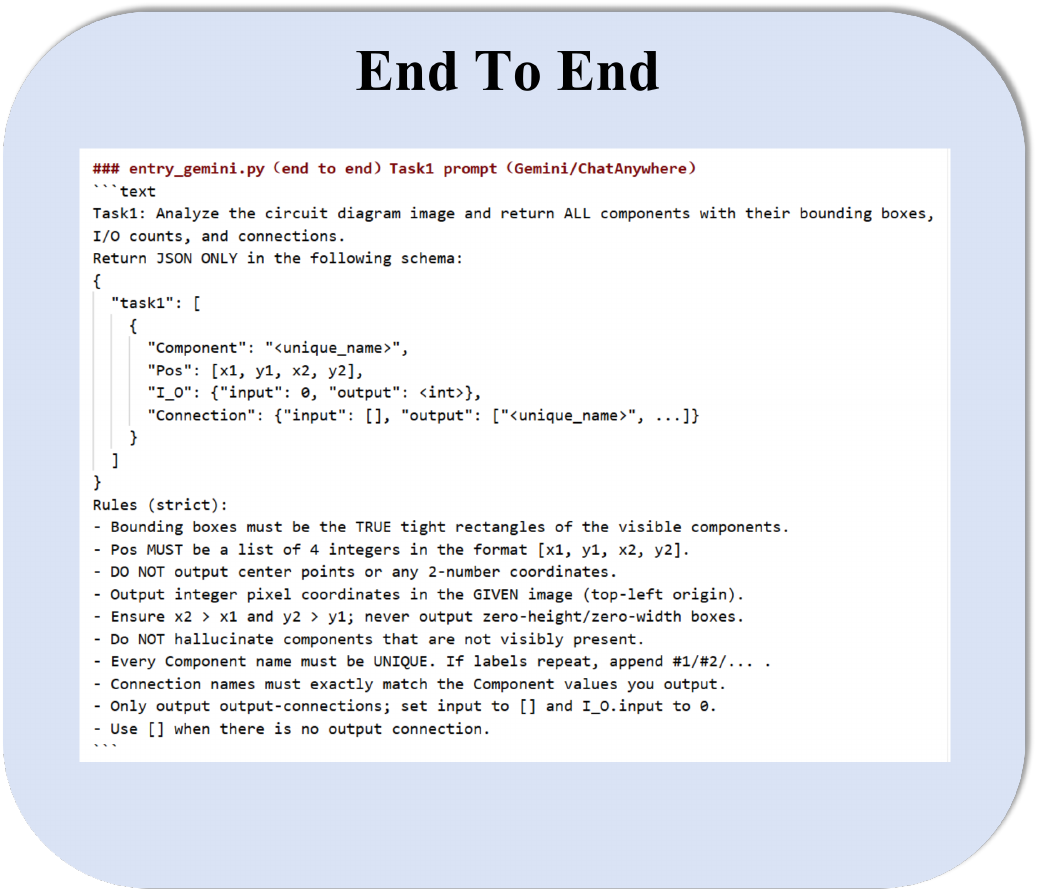}
        \caption{End-to-end prompt.}
        \label{fig:app-e2e-prompt}
    \end{subfigure}
    \hfill
    \begin{subfigure}[t]{0.32\textwidth}
        \centering
        \includegraphics[width=\linewidth]{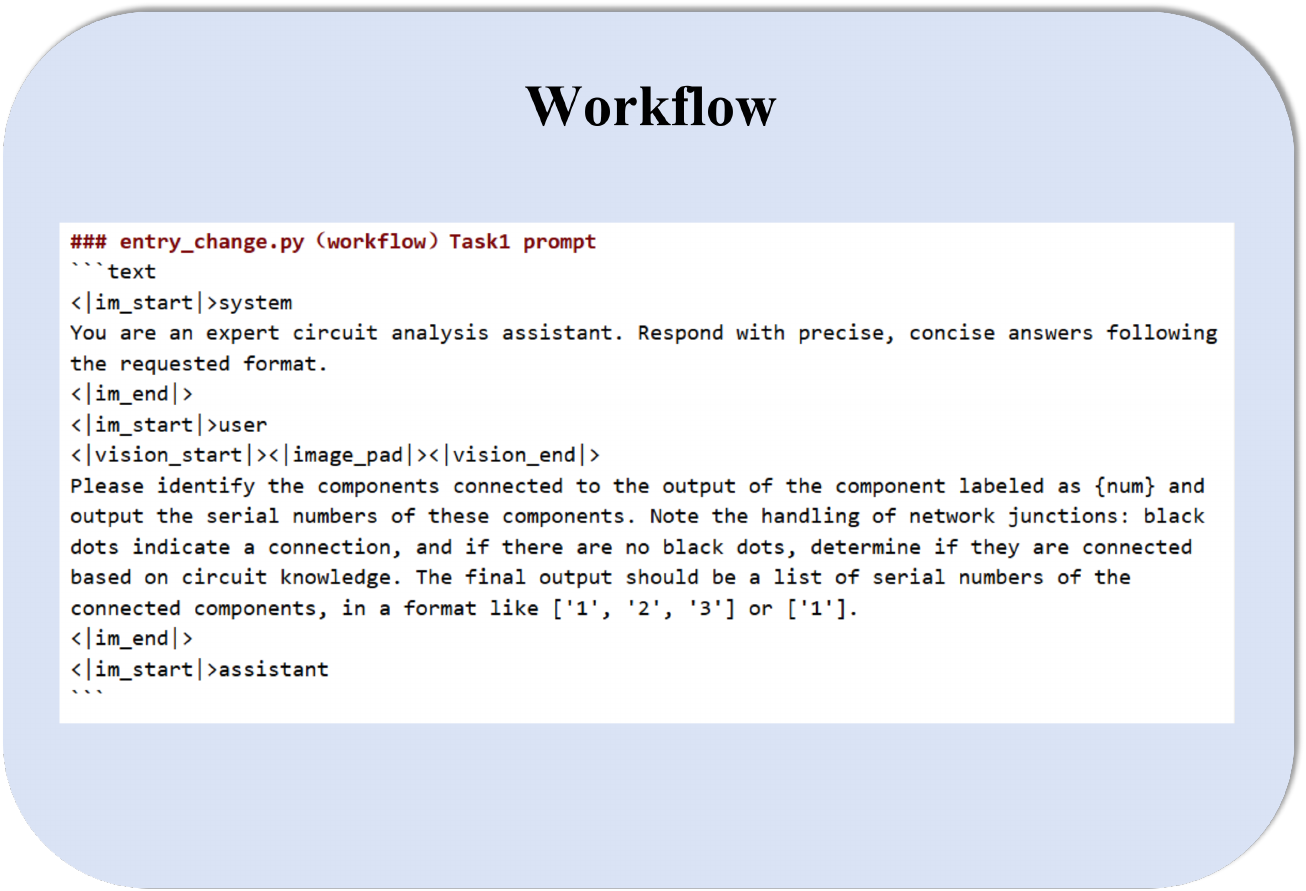}
        \caption{Workflow prompt.}
        \label{fig:app-workflow-prompt}
    \end{subfigure}
    \hfill
    \begin{subfigure}[t]{0.32\textwidth}
        \centering
        \includegraphics[width=\linewidth]{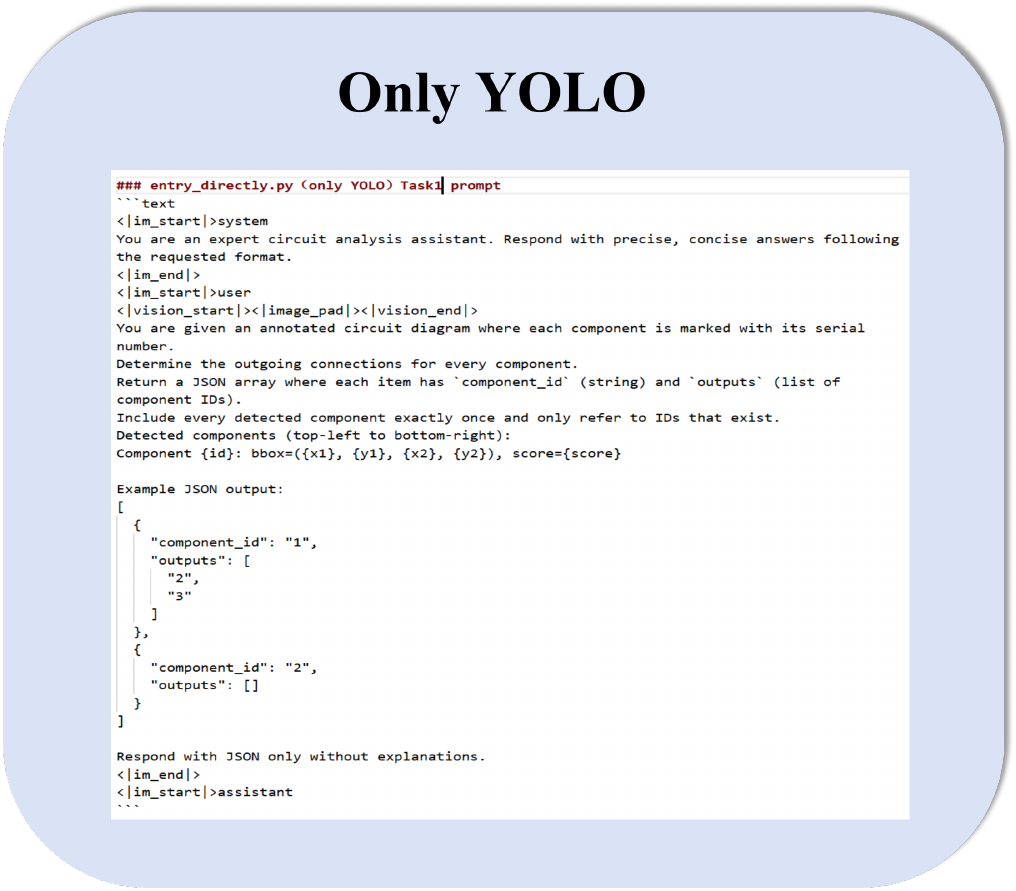}
        \caption{YOLO prompt.}
        \label{fig:app-yolo-prompt}
    \end{subfigure}
    \caption{Prompts used in Task Definitions.}
    \label{fig:app-prompts}
\end{figure}

To illustrate the rationale behind our method, this section presents the exploration of end-to-end, YOLO-assisted, and multi-agent workflow approaches for system-level diagram recognition, along with representative prompts shown in Figure~\ref{fig:app-prompts}.

Based on the task decomposition of DiagramNet into Listing, Localization, Connection, and Circuit QA, an end-to-end setting was first attempted (Figure~\ref{fig:app-e2e-prompt}), completing Task~1 (Listing, Localization, and Connection) in a single pass before directly answering Task~2. However, system-level diagrams often contain non-standard symbols, and their wiring is frequently expressed in implicit or crossing forms. As a result, preserving correctness for component recognition, localization, and topological connectivity simultaneously within one-round generation proves difficult. This limitation is reflected in the experimental results in Table~\ref{tab:overall-results}.

To address this limitation, YOLO annotations were introduced to strengthen Localization as the perception layer (Figure~\ref{fig:app-yolo-prompt}), enabling the model to observe detection boxes and indices. Yet requiring a single-pass output of the full connectivity graph still leads to instability because the non-deterministic YOLO detection order, combined with the non-unique mapping between indices and components, amplifies ambiguity and makes reliable connectivity recovery challenging.

These observations motivated the decoupled multi-agent workflow described in this paper (Figure~\ref{fig:app-workflow-prompt}). The Perception Agent first completes Listing and Localization, assigning unified indices to components in row-major order (left to right, top to bottom) to establish a stable sequential prior. The Reasoning Agent then predicts output connections component by component using $f_{\text{conn}}: I, c_i, \mathcal{C} \to \mathcal{T}_i$, which substantially reduces output-space complexity. Finally, Circuit QA is handled by the Knowledge Agent as a separate step. This workflow aligns with the DiagramNet task decomposition and significantly improves stability and accuracy for connectivity identification in complex topologies with dense wiring.

\section{Case Study}
\label{sec:app-case}

This appendix provides qualitative analysis of DiagramNet-3B and baseline methods across five aspects: end-to-end failures of commercial MLLMs on component detection, sources of S3 errors under visual ambiguity, domain knowledge gaps on textual QA, Netlistify failure modes, and cross-domain transfer to AMSBench.

\begin{figure}[ht]
    \centering
    \begin{subfigure}[t]{0.48\textwidth}
        \centering
        \includegraphics[width=\linewidth]{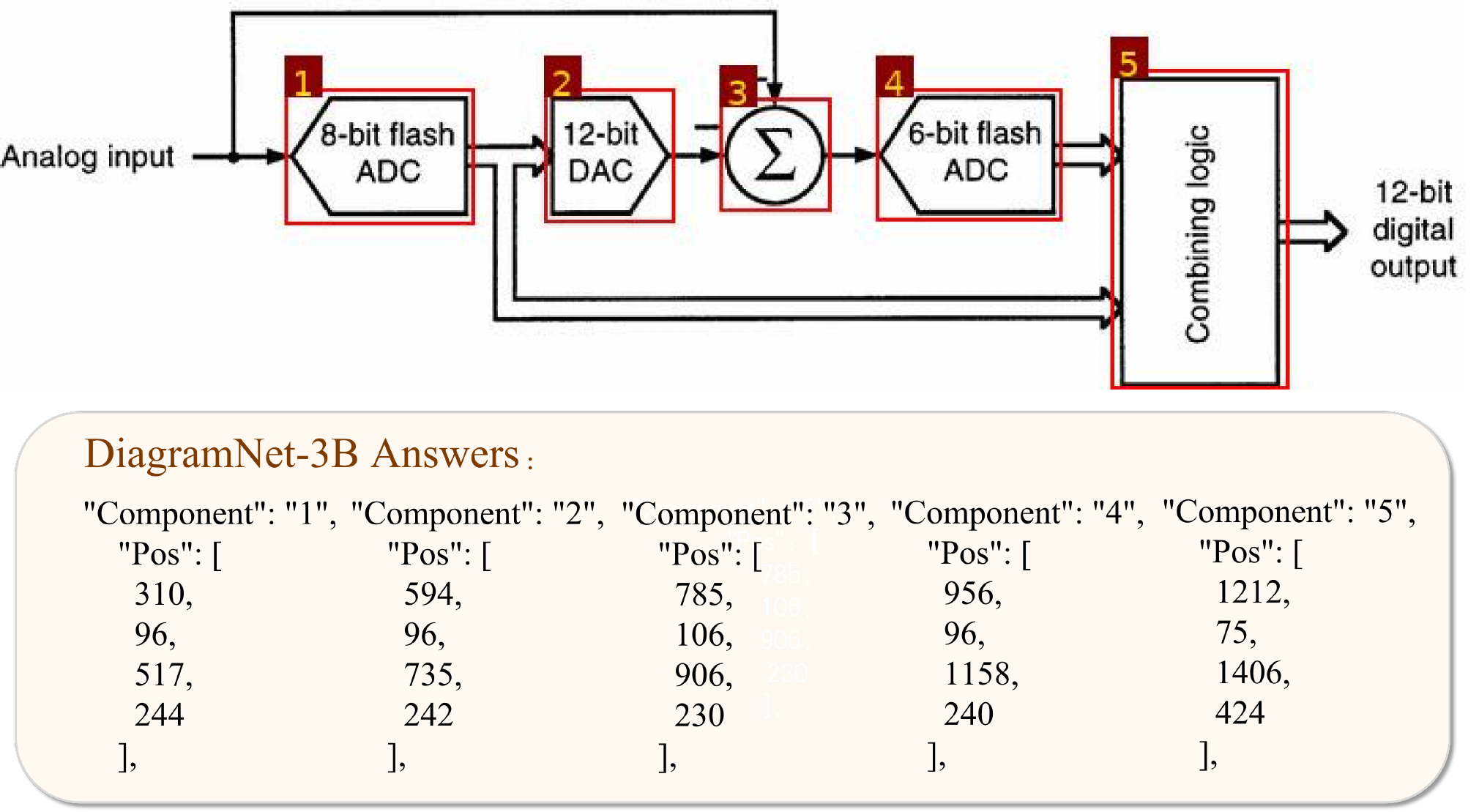}
        \caption{Workflow with YOLO.}
        \label{fig:mlmm-detect-a}
    \end{subfigure}
    \hfill
    \begin{subfigure}[t]{0.48\textwidth}
        \centering
        \includegraphics[width=\linewidth]{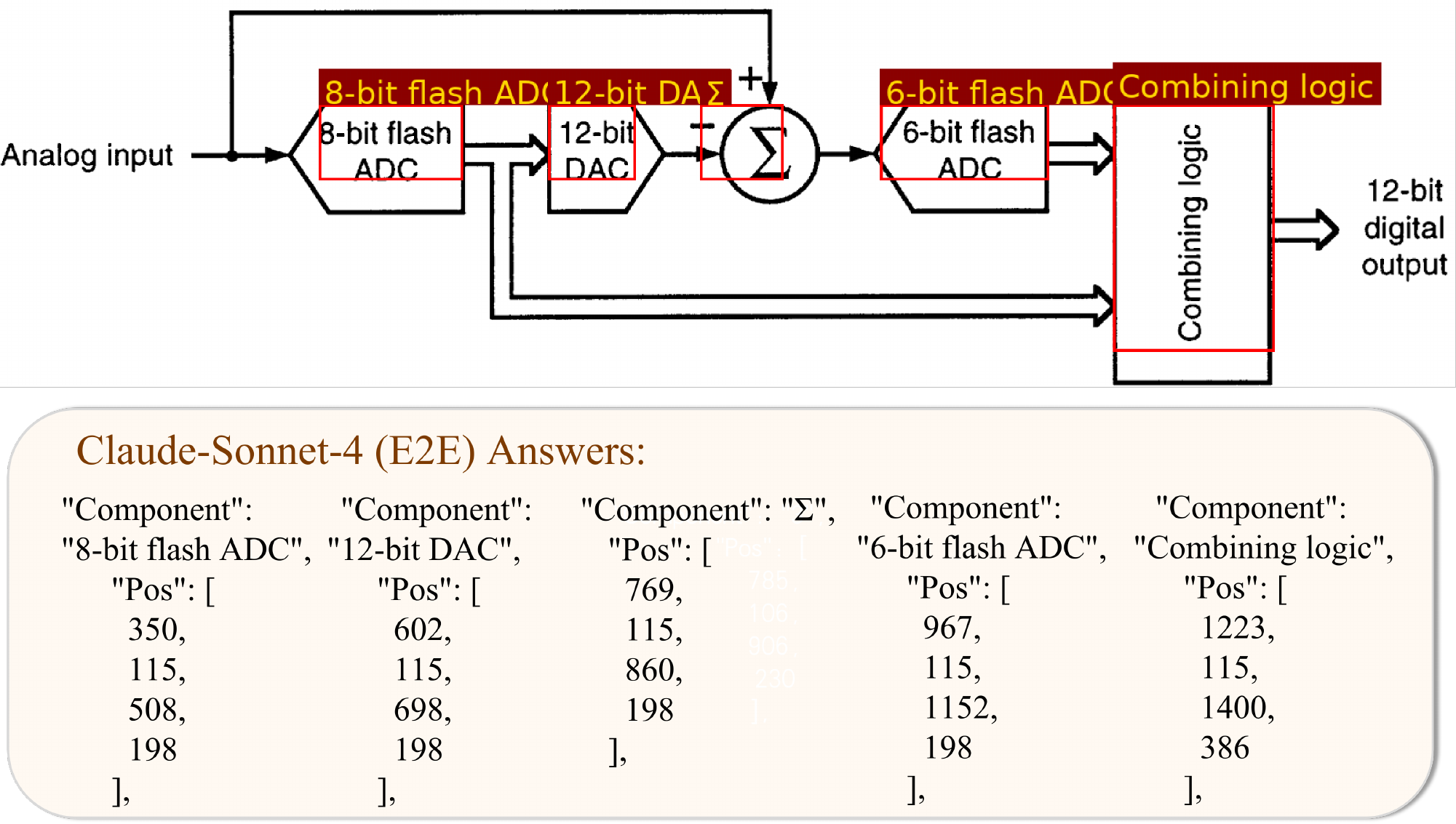}
        \caption{Claude Task~1 output.}
        \label{fig:mlmm-detect-b}
    \end{subfigure}

    \vspace{0.6em}
    \begin{subfigure}[t]{0.48\textwidth}
        \centering
        \includegraphics[width=\linewidth]{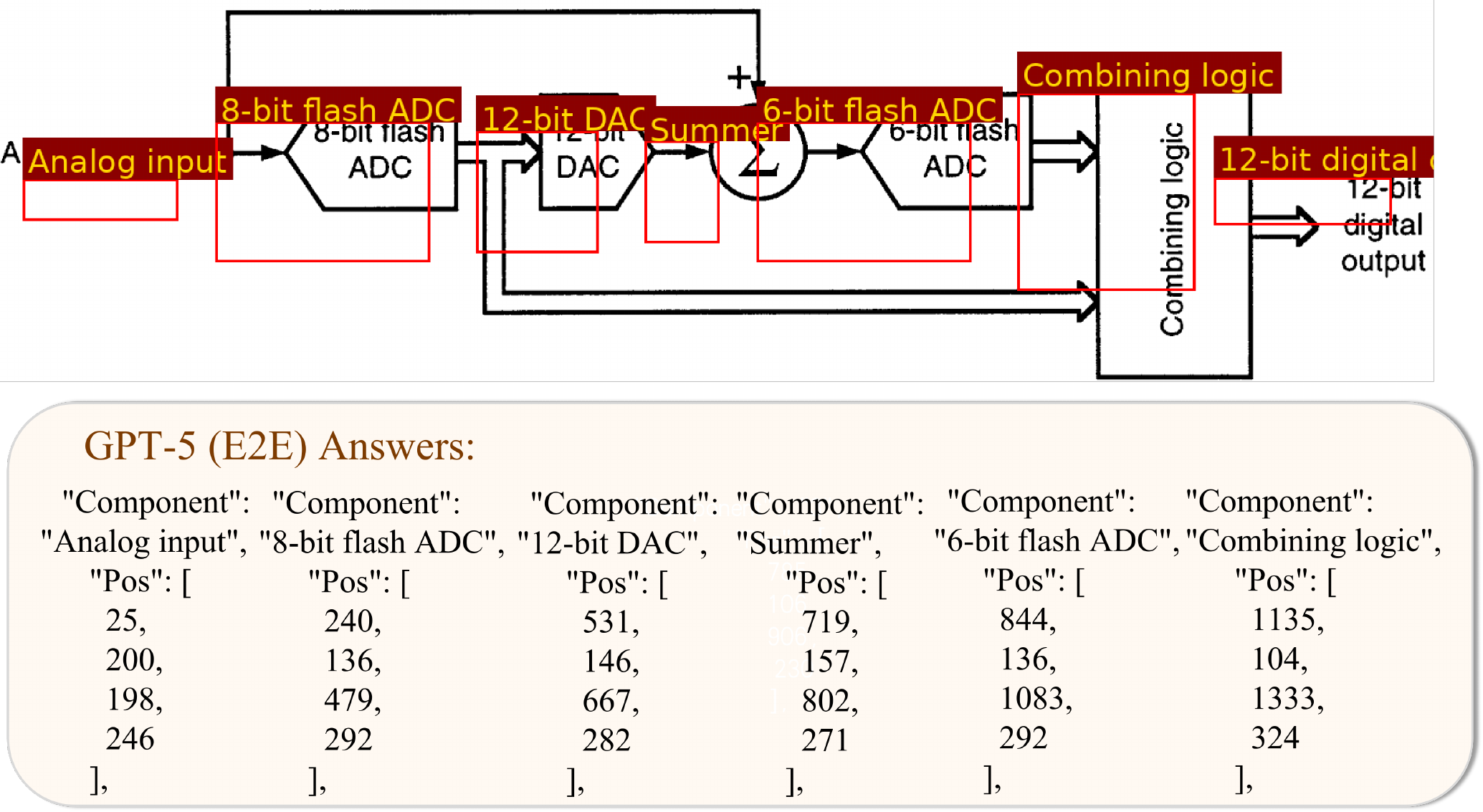}
        \caption{GPT-5 Task~1 output.}
        \label{fig:mlmm-detect-c}
    \end{subfigure}
    \hfill
    \begin{subfigure}[t]{0.48\textwidth}
        \centering
        \includegraphics[width=\linewidth]{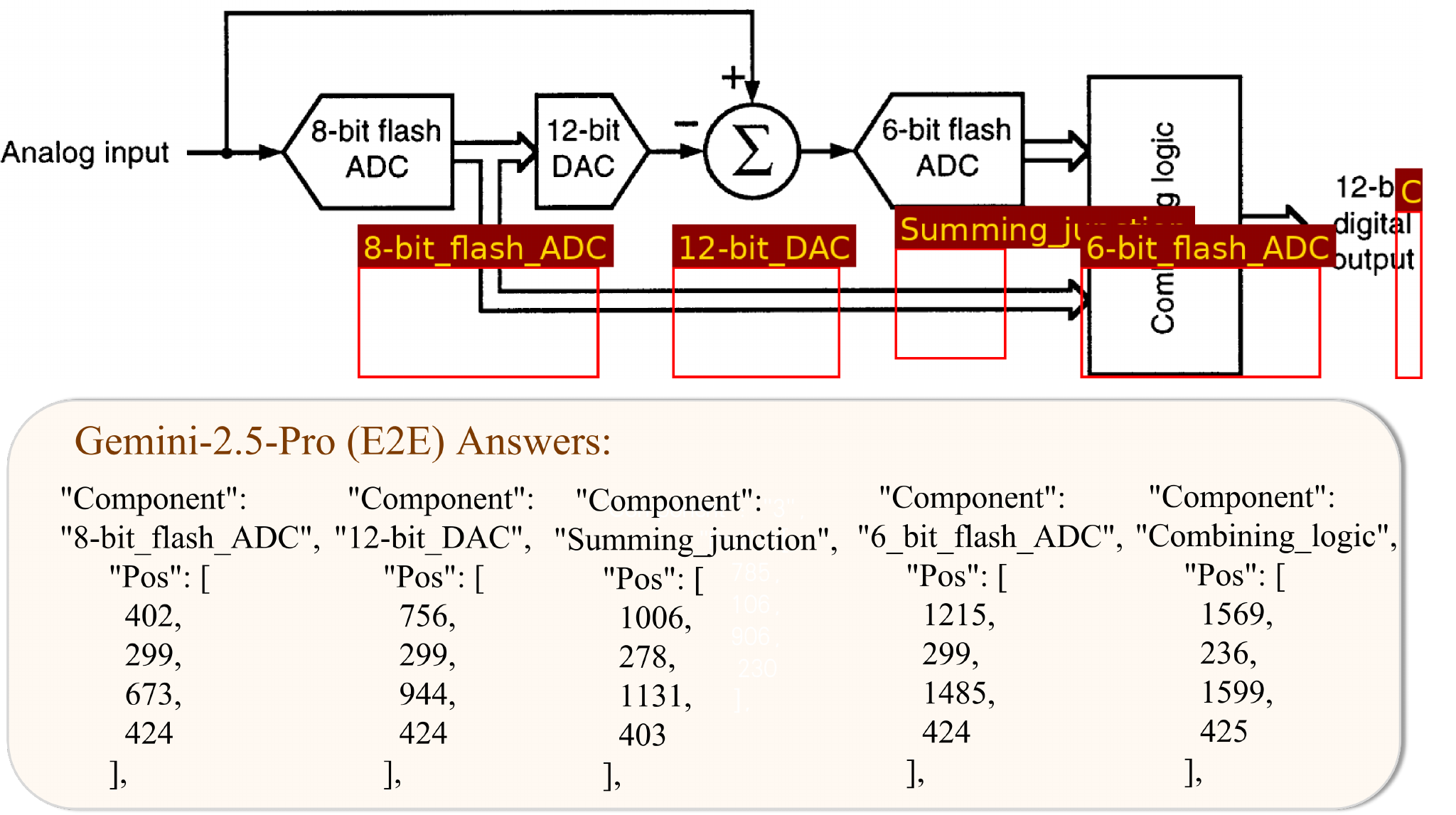}
        \caption{Gemini-2.5-Pro Task~1 output.}
        \label{fig:mlmm-detect-d}
    \end{subfigure}
    \caption{Component detection visualizations. (a) Our workflow with YOLO annotations. (b)--(d) Component locations rendered from Task~1 outputs of Claude, GPT-5, and Gemini-2.5-Pro.}
    \label{fig:mlmm-detect}
\end{figure}

\paragraph{End-to-end failures of commercial MLLMs on component detection.}
We evaluate GPT-5, Gemini-2.5-Pro, and Claude-Sonnet-4 on DiagramNet using identical prompts with strict output format constraints. All three models exhibit systematic failures on S1 (component detection): Gemini-2.5-Pro scores 0.008, GPT-5 scores 0.085, and Claude-Sonnet-4 scores 0.477. In contrast, our workflow achieves 0.988 using a perception agent.
Figure~\ref{fig:mlmm-detect} visualizes component detection results. Subfigure~(a) shows our workflow output with YOLO annotations. Subfigures~(b)--(d) show component locations rendered from Task~1 outputs of Claude, GPT-5, and Gemini-2.5-Pro.

We identify three failure patterns through visualization. \textbf{Invalid coordinates}: Gemini occasionally outputs bounding boxes with zero height, producing invalid localizations. This behavior is inconsistent across images, indicating unstable spatial alignment for certain diagram layouts. \textbf{Loose bounding boxes}: GPT-5 produces coordinates in reasonable ranges but with insufficient precision. The IoU frequently falls between 0.3 and 0.49, just below the S1 threshold. \textbf{Moderate accuracy}: Claude produces more stable coordinates with approximately 40\% hit rate, but this remains insufficient for high-precision detection.

After introducing our multi-agent workflow, Task~1 scores improve substantially: 128.7$\times$ for Gemini-2.5-Pro, 12.4$\times$ for GPT-5, and 1.7$\times$ for Claude-Sonnet-4. The dedicated detector provides reliable spatial references, and topology reasoning is decoupled from implicit localization. This demonstrates that the workflow mitigates the visual grounding bottleneck on circuit diagrams.

\paragraph{Failure mode analysis: sources of S3 errors on DiagramNet.}
\label{sec:s3-failure}
\begin{figure}[ht]
    \centering
    \begin{subfigure}[t]{0.32\textwidth}
        \centering
        \includegraphics[width=\linewidth]{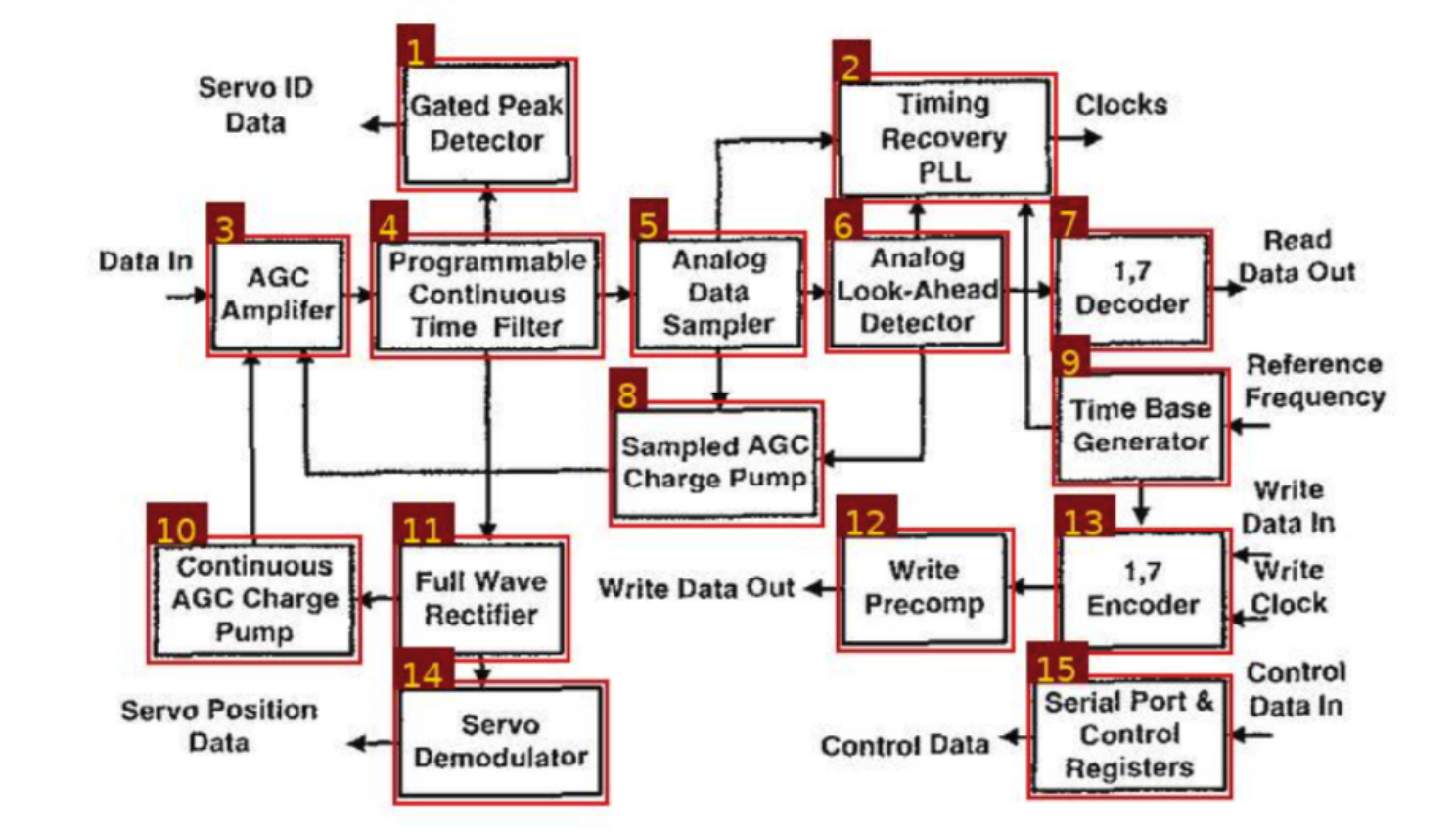}
        \caption{Weak directional cues.}
        \label{fig:s3-failure-a}
    \end{subfigure}
    \hfill
    \begin{subfigure}[t]{0.32\textwidth}
        \centering
        \includegraphics[width=\linewidth]{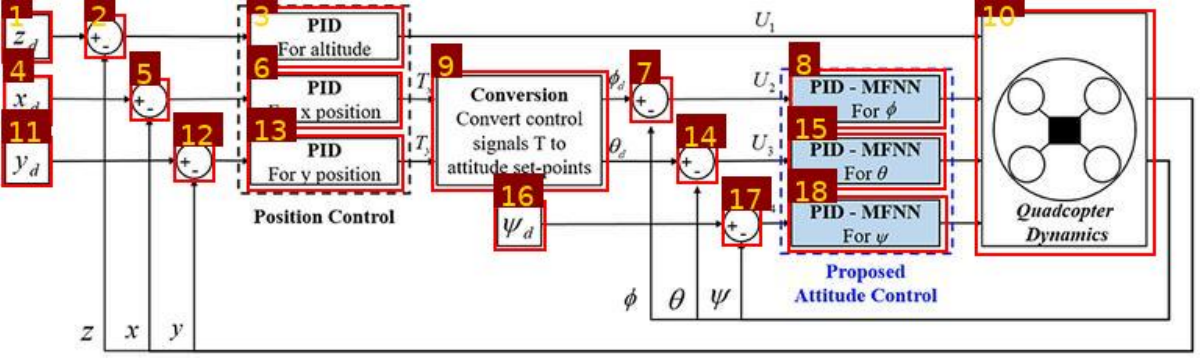}
        \caption{Crossing ambiguity.}
        \label{fig:s3-failure-b}
    \end{subfigure}
    \hfill
    \begin{subfigure}[t]{0.32\textwidth}
        \centering
        \includegraphics[width=\linewidth]{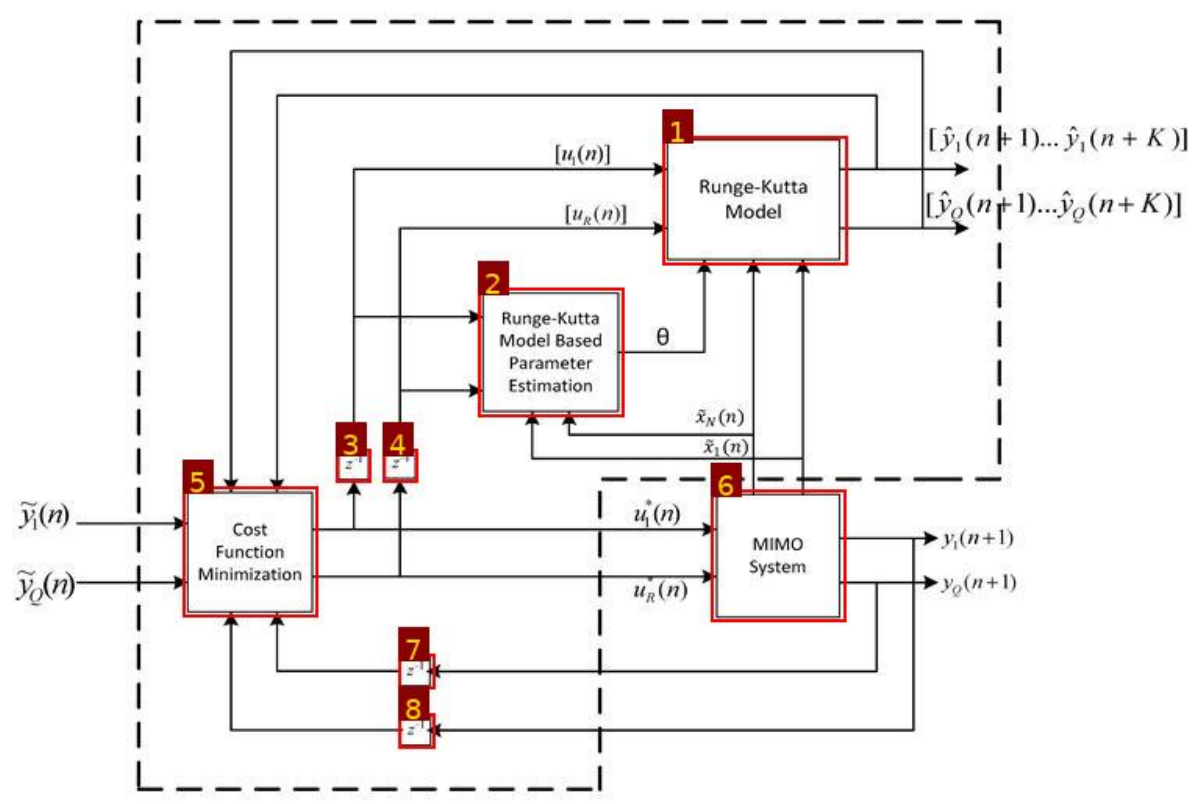}
        \caption{Multi-fan-out misses.}
        \label{fig:s3-failure-c}
    \end{subfigure}
    \caption{Representative S3 error cases on DiagramNet.}
    \label{fig:s3-failure}
\end{figure}
Our method achieves 0.855 on Task~1, compared to 0.862 for EDA Elite Winner. On S3, our score is 0.736 versus 0.777. The gap is concentrated in a small set of ambiguous samples and does not reflect a systematic failure of the recognition pipeline. Figure~\ref{fig:s3-failure} illustrates three triggering conditions. \textbf{Weak directional cues} cause direction ambiguity, introducing reversed edges. In Figure~\ref{fig:s3-failure-a}, unclear arrows swap the direction between device 3 and device 4. \textbf{Crossing ambiguity} arises when crossings lack explicit no-connection marks. The model may add false connections or miss true ones. In Figure~\ref{fig:s3-failure-b}, the output of device 7 is incorrectly linked to device 17. \textbf{Multi-fan-out structures} reduce endpoint visibility, causing missing branch targets. Figure~\ref{fig:s3-failure-c} shows missing edges under multi-fan-out. These errors correlate with limited visual cues and incomplete diagram markings. The primary direction for improvement is to enhance robustness under ambiguous conditions rather than to revise the overall pipeline.

\paragraph{Gap analysis on Textual QA.}
\begin{figure}[ht]
    \centering
    \includegraphics[width=0.65\textwidth]{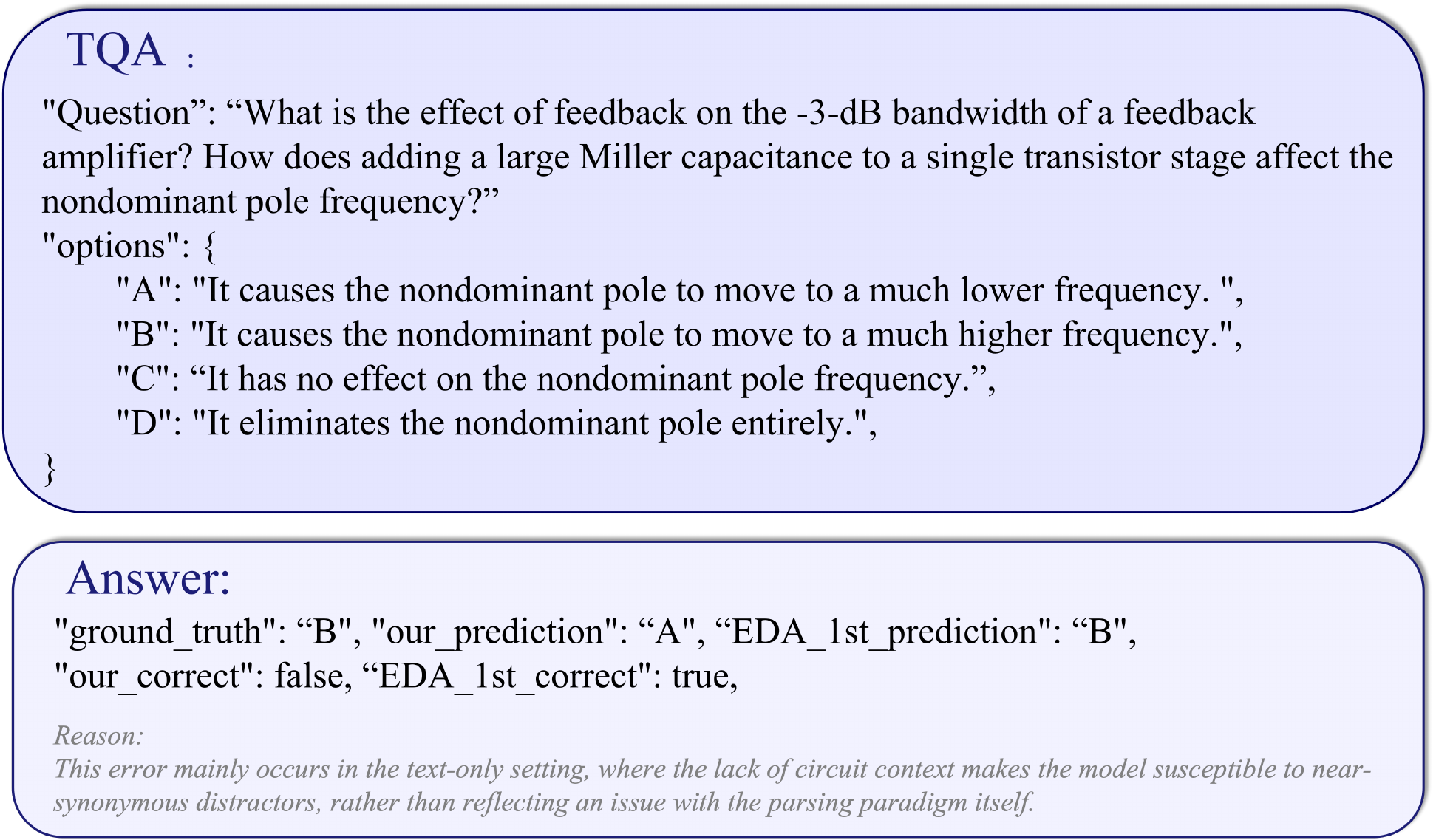}
    \caption{Gap analysis on the Textual Circuit QA subset of AMSBench.}
    \label{fig:tqa-gap}
\end{figure}

We compare DiagramNet-3B with EDA Elite Winner on the text-only TQA subset of AMSBench. On 30 questions, our model achieves 27/30 (accuracy 0.900), while EDA Elite Winner achieves 28/30 (accuracy 0.933). The two models differ on three questions; in two cases our model errs while the competitor answers correctly, explaining the net gap. The errors stem from fine-grained terminology distinctions in analog circuits. Specifically, our model confuses near-synonymous concepts that differ only by strict textbook definitions, such as MOS operating regions versus voltage-controlled resistors, and mispredicts the direction of nondominant pole shift under large Miller capacitance. The remaining gap reflects a coverage advantage on AMS-specific textbook facts rather than a reasoning deficiency. Both models answer graduate-level questions with short reasoning chains correctly, while errors concentrate on niche analog terminology. The comparison suggests that targeted text adaptation with concept-contrast supervision can close this gap.

\paragraph{Failure mode analysis: Netlistify on AMSBench and DiagramNet.}
Netlistify reports strong results on its original AMS benchmark but degrades on AMSBench and DiagramNet.
On AMSBench, Netlistify achieves F1=0.667 on Connection Judgement.
On DiagramNet S3, it scores only 0.15. We identify three failure modes.
\textbf{Symbol heterogeneity}: Netlistify's ResNet-based orientation classifier is trained on standard AMS symbols and fails to generalize to abstract system-level blocks with arbitrary shapes. \textbf{Text interference}: system-level diagrams contain dense annotations such as signal names and module labels. Netlistify's Transformer decoder misinterprets these text regions as connection endpoints, producing spurious edges. \textbf{Hyperparameter brittleness}: Netlistify requires manual tuning of distance thresholds and angle tolerances for each diagram style, whereas DiagramNet-3B learns these priors from data. These failure modes highlight the limitations of modular pipelines that separate visual feature extraction from topological reasoning.

\paragraph{Cross-domain transfer to AMSBench.}
We evaluate transfer learning by retraining the YOLO detector with 60 images from AMSBench. Because these images are used for detector training, we exclude component detection tasks and focus on two connection tasks and Textual QA to test connectivity reasoning and domain knowledge. On Connection Identification, DiagramNet-3B achieves F1=0.5, outperforming Netlistify (F1=0.433). Qualitative inspection shows that DiagramNet-3B transfers topological reasoning patterns from system-level diagrams to AMS circuits. For example, it correctly handles multi-terminal components such as operational amplifiers with five ports by reasoning over spatial layout, even though such structures are rare in the DiagramNet training set. DiagramNet-3B occasionally fails on AMS-specific conventions such as implicit ground connections or power rail shortcuts. This indicates that core topology reasoning transfers robustly while domain-specific conventions require targeted fine-tuning. Since AMSBench provides only test data without a training split, we leave this adaptation for future work.



\end{document}